\pdfoutput=1

\documentclass[11pt]{article}

\usepackage[]{acl}

\usepackage{times}
\usepackage{latexsym}

\usepackage{multirow}
\usepackage{amsmath}
\usepackage{graphicx}
\usepackage{placeins}
\usepackage{wrapfig}
\usepackage{bm}
\usepackage{booktabs}
\usepackage{makecell}
\usepackage{xcolor}
\usepackage{subfig}

\usepackage[T1]{fontenc}

\usepackage[utf8]{inputenc}

\usepackage{microtype}

\title{Adapting Task-Oriented Dialogue Models for Email Conversations}

\author{Soham Deshmukh \and Charles Lee \\ Microsoft, Redmond, USA \\ \texttt{\{sdeshmukh,charlle\}@microsoft.com}}

\begin{document}
\maketitle
\begin{abstract}
Intent detection is a key part of any Natural Language Understanding (NLU) system of a conversational assistant. Detecting the correct intent is essential yet difficult for email conversations where multiple directives and intents are present. In such settings, conversation context can become a key disambiguating factor for detecting the user's request from the assistant. One prominent way of incorporating context is modeling past conversation history like task-oriented dialogue models. However, the nature of email conversations (long form) restricts direct usage of the latest advances in task-oriented dialogue models. So in this paper, we provide an effective transfer learning framework (EMToD) that allows the latest development in dialogue models to be adapted for long-form conversations. We show that the proposed EMToD framework improves intent detection performance over pre-trained language models by 45\% and over pre-trained dialogue models by 30\% for task-oriented email conversations. Additionally, the modular nature of the proposed framework allows plug-and-play for any future developments in both pre-trained language and task-oriented dialogue models.
\end{abstract}
\section{Introduction}
\begin{figure*}
    \centering
    \includegraphics[width=6.3in]{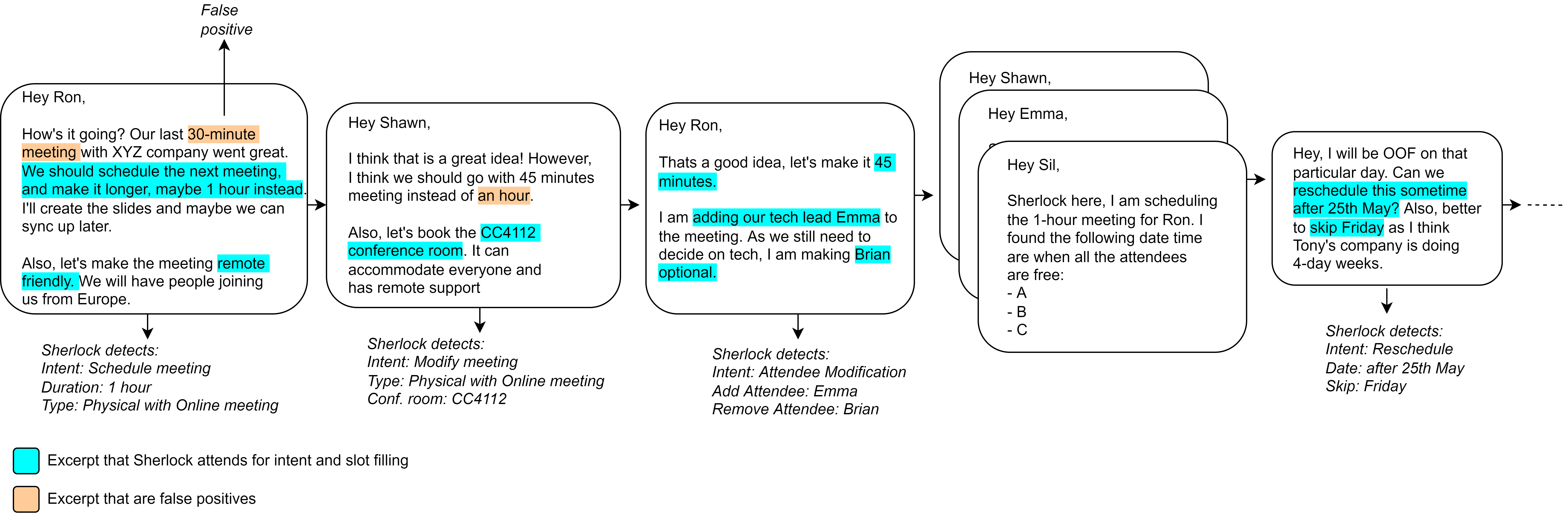}
    \caption{A simplified example of an email based assistant (Sherlock) for the task of scheduling meetings. In real life settings, the emails are longer with more diverse content not related to assistant's task of scheduling meetings}
    \label{fig:email assistant}
\end{figure*}

There has been an increase in task-oriented assistants working with long-form conversations like emails and documents. The backbone of such assistants is an NLU system \cite{7814394} which understands the intent of the user, extracts relevant entities, tracks the state of the conversation, and hence directs the conversation towards competition of the assistant's task. Such assistants allow better support of features like asynchronous and multiparty conversations and directives and bring a unique set of challenges. An example of such an email-based assistant is depicted in Figure \ref{fig:email assistant}. Some of the difficulties related to the lengthy nature of emails are tackled in recent works \cite{9003958, patra-etal-2020-scopeit, zhou2021narle, trajanovski-etal-2021-text}. 

However, building intent detection models for email or document conversations still has multiple challenges. First, the information necessary for the agent's task or user's ask from the agent can be buried in the email. The problem is made worse in multiparty conversations where user A is asking user B to do some task instead of the assistant. For example, User A asks User B to suggest alternate times instead of asking the assistant to find an alternate time for scheduling the meeting, here the correct intent detection by the assistant should be `no action'. This issue is coupled with false positives found in emails as shown in Figure \ref{fig:email assistant}.

Second, ambiguous utterances in the user's conversation. Continuing the last example, consider the example: "yes, all times work" as a new message in the conversation. Just purely from text, the assistant has no idea if this utterance is directed toward the assistant let alone the ask from the assistant. But if the assistant was aware of the past conversation or message, the assistant can disambiguate the utterance and detect the intent. Third, incorporating context has been shown to improve performance for intent detection and other tasks in both email and short conversation settings \cite{6639291, 10.1145/3331184.3331260, trajanovski-etal-2021-text}. The choice of context varies from task to task but the direction of incorporating context has led to interesting developments in pre-trained dialogue models \cite{wu-etal-2020-tod, zhang-etal-2020-dialogpt}. However, the latest developments in pre-trained dialogue models are focused on short-form conversations and focus on a small set of frequent phenomena \cite{naik2021adapting}. 

We aim to tackle the above three challenges by adapting task-oriented dialogue models. We introduce a simple yet effective framework called EMToD which enables building dialogue models for long-form conversations like emails. The framework consists of four main parts. First, extracting relevant sentences from user emails. Second, summarize the agent's email response by their actions. Third, modeling the last turn of the user and past history of the conversation. Fourth, aggregating and contextualizing the user's last message in conversation before final intent detection. Our contributions are summarized as follows:
\begin{itemize}
    \item We demonstrate that conversation context is essential for intent detection in task-oriented assistants working in multi-party settings with long conversations. We propose a framework to incorporate such context. 
    \item To the best of our knowledge, EMToD is the first transfer learning framework to adapt task-oriented dialogue models trained on short conversations to long-form conversations like emails. EMToD leverages pre-trained dialogue models for modeling conversations and adapts them in a simple and effective way for detecting intents in long-form conversations.
    \item We demonstrate how using EMToD results in large improvements to intent detection for email-based conversations over pre-trained language and dialogue models and analyze the effect of its individual components.
\end{itemize}

\begin{figure*}[!htb]
    \centering
    \includegraphics[width=0.9\textwidth]{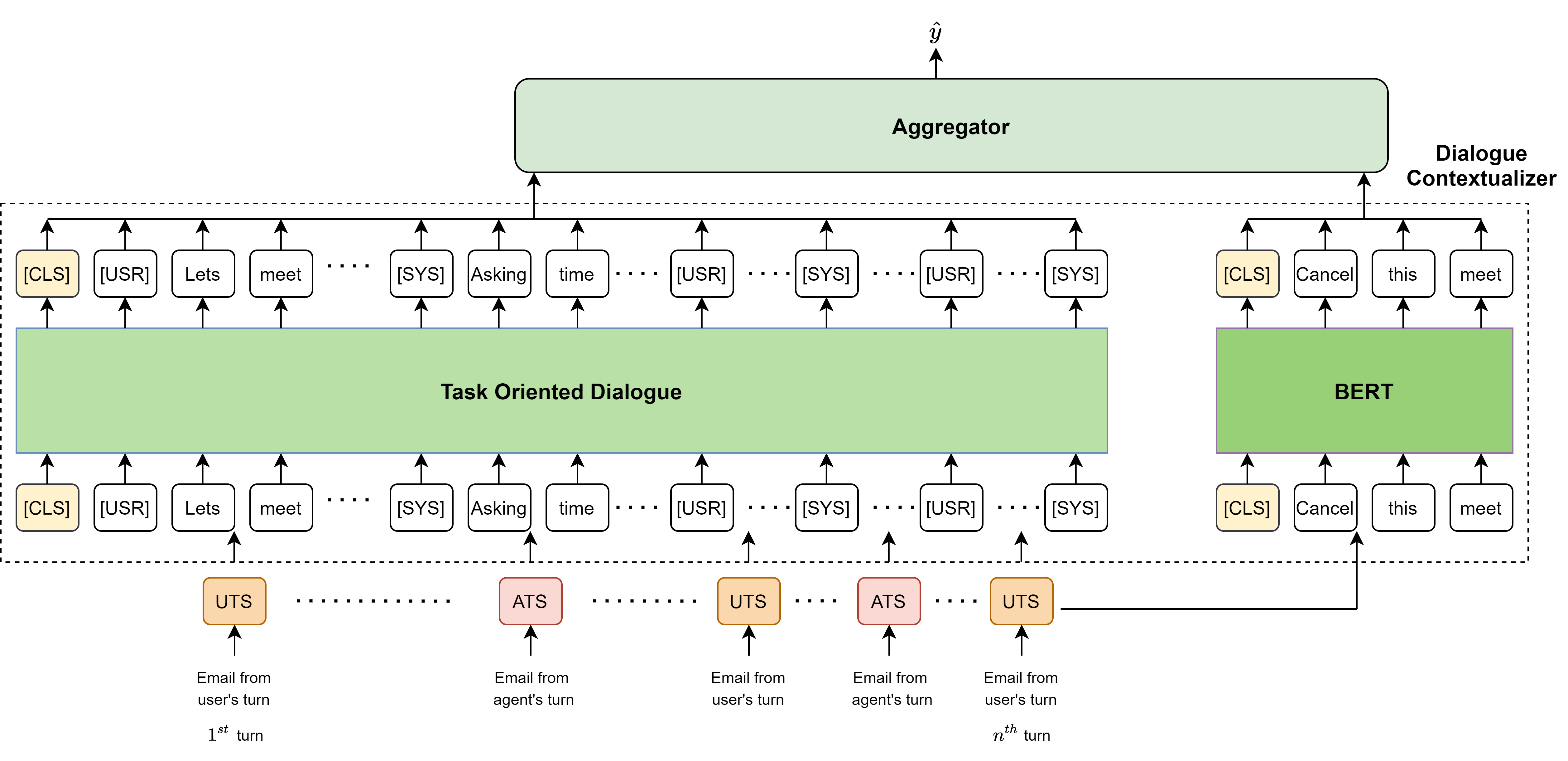}
    \caption{EMToD framework. Here UTS refers to `User text summarizer' and `ATS' refers to `Agent text summarizer'. The Dialogue Contextualizer incorporates both task-oriented dialogue model and language model.}
    \label{fig:EMToD}
\end{figure*}
\section{Framework details} \label{sec:framework_details}

In this section, we describe the notations used followed by the details of each module in the EMToD framework.

\textbf{Notations}. Let the dialogue between user and assistant be defined as $D = \{U_1, S_1, ... , U_t, S_t\}, t \in [0,N]$, where $U_t$ is the user's email at turn $t$, $S_t$ is the assistant's (system's) email at turn $t$ and $N$ is the total number of turns in the dialogue. $U_t$ can be from any user, specifically organizer or attendee in email settings due to multiparty nature of the conversation. The dialogue context at each turn $t$ is defined as $C_t = \{U_1, S_1, ... S_{t-1}, U_{t}\}$. Email at each turn from user ($U_t$) consists of multiple sentences ranging from 1 to m and is written as concatenation of all sentences ($U_{t_i}$) in the user turn. 
Similarly, each assistant's turn ($S_t$) can be decomposed into a concatenation of all sentences ($S_{t_i}$) in the assistant's turn. Let the summarized turn of user and assistant be denoted by $\hat{U}_i$ and $\hat{S}_i$ respectively.

\subsection{User turn summarizer} \label{sec:user_turn_summarizer}
The aim of the user turn summarizer is to summarize or extract relevant sentences to the assistant's task from user emails. We choose a particular form of extractive summarization approach called ScopeIt \cite{patra-etal-2020-scopeit}. ScopeIt consists of three parts: an intra-sentence aggregator, an inter-sentence aggregator, and a classifier to predict whether the given sentence is relevant to the assistant's task. The intra-sentence aggregator contextualizes the words within a sentence while the inter-sentence aggregator contextualizes the sentences within an email. 

The input to ScopeIt is the user's email $U_{t}$ at turn t. Let $U_{t_i} = \{w_{i,1} ... w_{i,l_i}\}$ be the $i^{th}$ sentence in $m$ total sentences, where $w_{j,k}$ denotes the $k^{th}$ word of the $j^{th}$ sentence. Similarly, each assistant's turn ($S_t$) can be decomposed into words occurring in the total sentences. 
Then the relevance of each sentence in user’s turn ($U_i$) is predicted as:
\begin{equation*}
\centering
    \begin{aligned}
        (e_{i,1}, \cdots e_{i, l_{i}}) &= BERT(w_{i,1} \cdots w_{i, l_i}) \\
    \end{aligned}
\end{equation*}

\begin{equation*}
\centering
    \begin{aligned}
    ({h_f}_{i,1} \cdots {h_f}_{i, l_i}) &= \overrightarrow{S2S}(e_{i,1}, \cdots e_{i, l_{i}}) \\
    ({h_b}_{i,1} \cdots {h_b}_{i, l_i}) &= \overleftarrow{S2S}(e_{i,1}, \cdots e_{i, l_{i}}) \\
    e_{s_i} &= [{h_f}_{i,l_i}; {h_b}_{i,1}] \\
    (f_{s_1}, \cdots f_{s_m}) &= \overleftrightarrow{S2S}(e_{s_1}, \cdots e_{s_m}) \\
    (p_{s_1}, \cdots p_{s_m}) &= (\sigma(f_{s_1}), \sigma(f_{s_2}) \cdots \sigma(f_{s_m}))
    \end{aligned}
\end{equation*}

Here BERT \cite{devlin-etal-2019-bert} is used for generating embeddings for each word \cite{patra-etal-2020-scopeit}. This is followed by a forward and backward S2S (Sequence to Sequence encoder) which adapts the embedding to the task. The concatenation of forwarding and backward embedding forms the sentence embedding $e_{s_i}$. The sentence embeddings are passed through another S2S to contextualize across email followed by a sigmoid function to determine the probability of the sentence being relevant to the task.  The relevant sentences determined by ScopeIt are then collected together to form the user's last turn ($\hat{U_t}$): 
$$\hat{U_t} = (U_{t_1},....U_{t_m}), \forall p_{t_m} >= \tau$$
where $p_{t_m}$ is the probability of the $m^{th}$ sentence in the $t^{th}$ turn being relevant and $\tau$ is a arbitrarily chosen threshold value.

\subsection{Agent turn summarizer} \label{sec:agent_turn_summarizer}
The aim of the agent turn summarizer is to summarize the agent's email. Similar to user turn summarizer, this can be a type of extractive summarizer \cite{patra-etal-2020-scopeit, liu2019finetune, zhong-etal-2020-extractive} or a heuristic. In task-oriented dialogue settings, the agent's email is about moving the conversation forward towards a completion state by either asking questions, reporting status, providing acknowledgments, etc. The agent's response can be completely or part naturally generated but is about filling a particular slot value missing to complete the task. This allows the usage of simple regex-based or classification models as a substitute for larger extractive summarizer models. The summary heuristic is inspired by action-based annotation \cite{10.5555/3504035.3504630} whereas in our case the summary is about the action taken by the email assistant. Hence, the summary of the agent's turn ($\hat{S}_t$) can be a concatenation of multiple sentences from an extractive summarizer or an action predicted through a regex pattern matching or text classifier. Then the relevant sentences determined by the User and Agent turn summarizer together form the conversation history ($\hat{C}_t$): 

$$\hat{C}_t = \{\hat{U}_1, \hat{S}_1, ... \hat{S}_{t-1}, \hat{U}_t\}$$

\subsection{Dialogue Contextualizer} \label{sec:dialog_contextualizer}
The dialogue contextualizer module consists of a pre-trained dialogue model which models and adapts the conversation history ($C_t$) toward the agent's task. For modeling conversation, any latest pre-trained task-oriented dialogue encoder can be used \cite{zhang-etal-2020-dialogpt, wu-etal-2020-tod, hosseiniasl2020simple}. However, the user's last turn might be very small compared to the past conversation history of multiple turns. This might lead to the intent of the past turn or largest turn (in terms of character length) being predicted. To overcome this problem, we also encode the user's last turn text separately using pre-trained language model \cite{devlin-etal-2019-bert, sanh2020distilbert, Liu2019RoBERTaAR, clark2020electra, he2021deberta}. So the contextualizer returns two types of embeddings: conversation and user's last turn embeddings separately. Each embedding consists of the embedding of CLS ($X_{CLS}$) token and word tokens ($X_{tokens}$)
\begin{equation}
    \begin{aligned}
        X_{1_{CLS}}, X_{1_{tokens}} = DIALOG(\hat{C}_t) \\ \nonumber
        X_{2_{CLS}}, X_{2_{tokens}} = BERT(\hat{U}_t)
    \end{aligned}
\end{equation}

\subsection{Aggregator} \label{section_aggregator}
The aggregator combines the conversation embedding and the user's last turn text embedding to produce a final embedding. This final embedding will be used by a classifier to detect intents. We explore three methods of aggregation (fusion): concatenation, attention \cite{NIPS2017_3f5ee243}, and cross attention. 

\textbf{Concatenation. } The direct way to fuse the two embeddings is the concatenation of the $CLS$ tokens from both embeddings. The $CLS$ token from the pre-trained dialogue and language model is a global feature representation of the conversation history and user's last turn respectively. Also, concatenating $CLS$ tokens reduces processing to only one token. The concatenated $CLS$ tokens are then passed to a classifier or linear layer and Sigmoid function ($f(.)$) to get a final prediction. 
$$ \hat{y} = f(X_{1_{CLS}}, X_{2_{CLS}})$$

\textbf{Attention. } We use variant of attention called scaled dot product attention \cite{NIPS2017_3f5ee243}.
In our setup, the query is the user's last turn $CLS$ embedding ($X_{1_{CLS}}$), and the key and value are the conversation's token embeddings ($X_{2_{tokens}}$) obtained through dialogue model. This allows for contextualizing the user's last turn information with respect to the past conversation and allows to keep only the relevant tokens from the past conversation history. The resulting vector is then passed to a linear layer and Sigmoid function ($f(.)$) to get a final prediction. 
$$ \hat{y} \text{ = }f(\text{softmax}(\frac{X_{1_{CLS}}X_{2_{tokens}}^T}{\sqrt{d_k}})X_{2_{tokens}})$$

\textbf{Cross attention. } This extends the previous scaled dot product attention mechanism. Rather than only using the last turn $CLS$ token as a query to contextualize past conversation history, it also uses the conversation history $CLS$ token to contextualize the user's last turn. This ensures that the information has been exchanged both ways and the tokens of the last user turn are appropriately weighted. Then the final two embeddings are concatenated followed by a linear layer and Sigmoid function ($f(.)$) to get the final prediction.

\begin{minipage}{0.42\textwidth}
    \centering
        \begin{equation*}
            \begin{aligned}
            y_1 \text{ = softmax}(\frac{X_{1_{CLS}}X_{2_{tokens}}^T}{\sqrt{d_k}})X_{2_{tokens}} \\
            y_2 \text{ = softmax}(\frac{X_{2_{CLS}}X_{1_{tokens}}^T}{\sqrt{d_k}})X_{1_{tokens}}
            \end{aligned}
            \end{equation*}
\end{minipage}

\begin{minipage}{0.45\textwidth}
        \centering
        \begin{equation*}
            \begin{aligned}
            \hat{y} = f(y_1, y_2)
            \end{aligned}
    \end{equation*}
\end{minipage}

The entire model shown in figure \ref{fig:EMToD} is end-to-end trained using binary cross entropy loss between predicted intents ($\hat{y}$) and ground truth intents ($y$). We leave exploration of different attention mechanism as future work \cite{deshmukh2021improving,deshmukh2020multi}.

\begin{table}[]
\small
\begin{tabular}{c|c|c|ccc}
\hline
\multirow{2}{*}{Dataset} & \multirow{2}{*}{\makecell{Dialo\\gues}} & \multirow{2}{*}{\makecell{User \\ utteran.}} & \multicolumn{3}{c}{Statistics} \\
& & & Mean & Std. & Max \\ \hline
Train/Val & 56,462 & 123,468 & 4.49 & 4.84 & 119.0 \\
Test & 10,502 & 23,526 & 4.53 & 4.92 & 104.0 \\ \hline
\end{tabular}
\caption{\label{table: dataset}
Dataset description
}
\end{table}

\begin{table*}[]
\centering
\small
\begin{tabular}{ccccclllll} 
\hline
\multicolumn{5}{c}{\textbf{Model}} & \multicolumn{5}{c}{\textbf{Metrics}} \\
dialogue & email & aggregator & user & agent & \multicolumn{1}{c}{micro-F1} & \multicolumn{1}{c}{macro-F1} & \multicolumn{1}{c}{micro-P} & \multicolumn{1}{c}{macro-P} & \multicolumn{1}{c}{acc.} \\ \hline
- & BERT & - & - & - & 0.644 & 0.232 & 0.536 & 0.208 & 0.370 \\
- & BERT & - & ScopeIt & - & 0.754 & 0.362 & 0.679 & 0.315 & 0.500 \\
ToD-BERT & - & - & - & - & 0.718 & 0.319 & 0.687 & 0.311 & 0.427 \\
ToD-BERT & - & - & ScopeIt & - & 0.779 & 0.361 & 0.771 & 0.369 & 0.523 \\
ToD-BERT & - & - & ScopeIt & Trunc. & 0.769 & 0.374 & 0.770 & 0.379 & 0.516 \\
ToD-BERT & - & - & ScopeIt & Summar. & 0.824 & 0.457 & 0.824 & 0.451 & 0.597 \\
ToD-BERT & BERT & Concat & ScopeIt & Summar. & 0.905 & 0.614 & 0.904 & 0.592 & 0.773 \\
ToD-BERT & BERT & Attn. & ScopeIt & Summar. & 0.893 & 0.672 & 0.884 & 0.629 & 0.756 \\
ToD-BERT & BERT & Cross-Attn. & ScopeIt & Summar. & \textbf{0.938} & \textbf{0.746} & \textbf{0.937} & \textbf{0.734} & \textbf{0.846} \\
\hline
\end{tabular}
\caption{\label{table: main results}
Intent detection results from EMToD experiments. Here - indicates absense of the particular module.
}
\end{table*}
\section{Experiments}

\subsection{Dataset creation}
The dataset is built by sampling dialogues from live multi-party user interactions with an email-based assistant for the task of scheduling meetings. This email-based assistant is a consumer service that allows eyes-on access to user data for restricted individuals under privacy regulations. The user utterances consist of multiple users participating in the email thread asking about scheduling, modifying meeting parameters, and canceling meetings along with the normal (not related to the agent's task) conversation going on with other attendees. Both the users and assistants converse in English. The dataset schema consists of a total of 13 intents. 

The training data consists of 314,970 dialogues from January 2021 to May 2021. We filter the dialogues to remove threading errors and bugs. This results in 56,462 dialogues (123,468 user utterances) in the training dataset, out of which 10\% is allocated for validation. The conversation from May 2021 to July 2021 are used as unseen test data. This consists of 77,612 conversations and is passed through similar filtering as train data resulting in 10,502 dialogues (23,526 user utterances). The train and test data statistics are described in Table \ref{table: dataset}. We perform data augmentation in user utterances by changing user asks and modifications to meetings like mode of meeting (online vs conference call), the mode of online meeting (Teams, Zoom, Skype, etc), location of the meeting, duration, timezone, etc. 

\subsection{Baselines} \label{sec:baselines}
The baselines consist of language models, dialogue models, and their combination for intent detection.

\textbf{Language model.} The language model baseline consists of pre-trained: BERT \cite{devlin-etal-2019-bert}, lite BERT variants \cite{sanh2020distilbert, conf/iclr/LanCGGSS20} and latest advances in transformer pre-training and architectures \cite{Liu2019RoBERTaAR, clark2020electra, he2021deberta}.

\textbf{Dialogue model.} There has been an increase in pre-trained dialog models \cite{wu-etal-2020-tod, zhang-etal-2020-dialogpt, hosseiniasl2020simple}. We choose ToD-BERT as the baseline and further backbone of EMToD. This is due to two reasons: first, the ToD-BERT model is built for task-oriented dialogue for specific domains similar to how the EMToD framework will be used. Second, the training dataset used for ToD-BERT consists of 8 domain-specific task-oriented dialogue datasets including calendar reservation. This makes ToD-BERT a suitable pre-trained baseline model in various domains. 

\textbf{Summarization. } We start with no summarization for both the user's and the agent's turn as a baseline. For the user's turn we specifically use ScopeIt due to its proven benefits for email summarization in task-oriented settings \cite{patra-etal-2020-scopeit}. With respect to the agent's response summarization, we use direct truncation to a specified length (trunc.) and regex heuristic to summarise the agent's response into 35 unique actions or intents.

\textbf{Aggregator. } We compare three aggregator methods: concatenation, attention, cross attention. Refer section \ref{section_aggregator} for the aggregator details.

\subsection{Experimental setup}
All the models are trained on 4 Tesla K80 GPUs (12 GB each) using data parallelism with a batch size of 128 and a learning rate of 1e-3. In all the experiments, we use a maximum of 30 epochs with early stopping based on performance on the validation set. The gradients are not propagated through pre-trained language and dialog models and only the subsequent modules are trained due to computing constraints. The implementation of all the pre-trained language, dialogue models, and tokenizers are based on the HuggingFace Transformers library \cite{wolf2020huggingfaces}. The baselines in Section \ref{sec:baselines} and \ref{sec: main results} extract CLS token from the pre-trained models followed by a linear classifier to predict intents.

\section{Results}

\begin{table*}[]
\centering
\small
\begin{tabular}{cccllllll} 
\hline
\multicolumn{3}{c}{\textbf{Model}} & \multicolumn{5}{c}{\textbf{Metrics}} \\
dialogue & email & aggregator & \multicolumn{1}{c}{micro-F1} & \multicolumn{1}{c}{macro-F1} & \multicolumn{1}{c}{micro-P} & \multicolumn{1}{c}{macro-P} & \multicolumn{1}{c}{acc.} & \multicolumn{1}{c}{params.}\\ \hline
ToD-BERT & BERT & Cross Attn. & 0.938 & 0.746 & 0.937 & 0.734 & 0.846 & 288.1 \\
ToD-DistilBERT & DistilBERT & Attn. & 0.888 & 0.619 & 0.883 & 0.592 & 0.751 & 201.6\\
ToD-DistilBERT & DistilBERT & Cross Attn. & 0.937 & 0.744 & 0.932 & 0.716 & 0.844 & 202.0\\
ToD-DistilBERT & ALBERT & Attn. & 0.885 & 0.6032 & 0.895 & 0.580 & 0.746 & 78.5\\
ToD-DistilBERT & ALBERT & Cross Attn. & 0.931 & 0.706 & 0.927 & 0.682 & 0.824 & 78.9\\
 \hline
\end{tabular}
\caption{\label{table: small models}
Effect of distilled models on EMToD performance
}
\end{table*}

\begin{table*}[]
\centering
\small
\begin{tabular}{cccllllll} 
\hline
\multicolumn{3}{c}{\textbf{Model}} & \multicolumn{5}{c}{\textbf{Metrics}} \\
dialogue & email & aggregator & \multicolumn{1}{c}{micro-F1} & \multicolumn{1}{c}{macro-F1} & \multicolumn{1}{c}{micro-P} & \multicolumn{1}{c}{macro-P} & \multicolumn{1}{c}{acc.} & \multicolumn{1}{c}{params.}\\ \hline
ToD-BERT & BERT & Cross Attn. & 0.938 & 0.746 & 0.937 & 0.734 & 0.846 & 288.1\\
ToD-DistilBERT & DistilBERT & Cross Attn. & 0.937 & 0.744 & 0.932 & 0.716 & 0.844 & 201.6\\
ToD-DistilBERT & RoBERTa & Cross Attn. & 0.928 & 0.684 & 0.919 & 0.651 & 0.820 & 191.9\\
ToD-DistilBERT & ELECTRA & Cross Attn. & 0.919 & 0.634 & 0.917 & 0.620 & 0.791 & 176.1\\
ToD-DistilBERT & DeBERTa & Cross Attn. & 0.910 & 0.626 & 0.893 & 0.596 & 0.791 & 205.8\\
 \hline
\end{tabular}
\caption{\label{table: robust models}
Effect of robust transformer models for user turn modelling on EMToD performance.
}
\end{table*}

\subsection{Main results} \label{sec: main results}
We first benchmark two main baselines on the dataset: pre-trained language model \cite{devlin-etal-2019-bert} and pre-trained dialogue model \cite{wu-etal-2020-tod}. As ScopeIt \cite{patra-etal-2020-scopeit} is shown to improve performance on other downstream tasks, we use a combination of ScopeIt and pre-trained models as our third set of baseline. The results are shown in Table \ref{table: main results} and averaged across 3 runs. The dialogue column and the email column refer to the dialogue and text encoder used to embed the conversation history and user's last turn text respectively. Using pre-trained dialogue model (ToD-BERT) compared to pre-trained language model (BERT) leads to a micro F1 improvement of about 11\%. The addition of ScopeIt to both BERT and ToD-BERT leads to about 17\% and 8\% respectively. This shows the importance of detecting and filtering out relevant sentences to the agent's task from user emails. The agent response emails are not summarized so far till row 4 in Table \ref{table: main results}.

We introduce simple truncation (Trunc.) of the agent's email response to arbitrary length decided based on data statistics. This keeps the metrics more or less the same across different metrics indicating the need for a more intelligent way of truncating or extracting information from the agent's email. Then we introduce the agent's turn summarization as described in section \ref{sec:agent_turn_summarizer} and section \ref{sec:baselines}. By intelligently summarising the agent's turn (Summar.) improves micro F1 score by 5\% over ToD-BERT with ScopeIt. To capture the user's last turn better, we introduce a separate text encoder (BERT) and evaluate different aggregation methods. The additional separate text encoder leads to an improvement in at least 8\%. The best performance obtained with EMToD framework is with ToD-BERT and BERT for dialogue and email encoder respectively followed by cross-scaled dot product attention (cross attn.) aggregation function. The EMToD framework leads to about 45\% and 30\% improvement in F1 micro scores over pre-trained language and dialogue models respectively. 

\subsection{Effects of using distilled models} \label{sec: results from distil models}
We evaluate the effect of using distilled and models with fewer parameters \cite{sanh2020distilbert, wu-etal-2020-tod, conf/iclr/LanCGGSS20}. The distilled models have been shown to provide similar performance but at a fraction of training time, inference latency, and model size.  For this experiment, we start with the best ToDTL configuration with ScopeIt as user turn summarizer and Heuristics to summarize agent's responses as a baseline as shown in Table \ref{table: small models}. The number of parameters (params.) reported is in Millions. Using smaller models like BERT and ALBERT does reduce performance measured by micro-F1. The choice of aggregator has more impact where cross attention leads to about 5.5\% improvement over-scaled dot product attention. The EMToD configuration dialogue encoder as ToD-DistilBERT, email encoder as DistilBERT, and aggregator as Cross Attention have negligible performance drop and reduce per epoch training time by 30\%. The inference latency is studied in Section \ref{sec: Inference time}.

\subsection{Effect of robust transformer models}
We also study the effect of improved pre-trained language models for user turn on EMToD performance. For this, we use recent developments on top of BERT using better optimization strategies, larger training data, and newer learning methods \cite{Liu2019RoBERTaAR, he2021deberta, clark2020electra}. We use the best performing configuration from Table \ref{table: main results} of EMToD as the baseline. From Table \ref{sec: results from distil models} we can see the improved language models (RoBERTa, ELECTRA, DeBERTa) do not result in better performance for intent detection. From this, we can infer that though EMToD relies on last user turn embeddings for detecting intents, it is not the major contributing factor in determining the intents.

\begin{table}[]
\centering
\small
\begin{tabular}{@{}lll@{}}
\hline
\multicolumn{1}{c}{\multirow{2}{*}{\textbf{System}}} & \multicolumn{2}{c}{\textbf{Accuracy}} \\ \cmidrule(l){2-3} 
\multicolumn{1}{c}{} & \multicolumn{1}{c}{week 1} & \multicolumn{1}{c}{week 2} \\ \hline
Internal & 0.873 & 0.878 \\
EMToD & 0.924 & 0.927 \\ \hline
\end{tabular}
\caption{\label{table: deployment results} 
EMToD against internal system on live data
}
\end{table}

\begin{table*}[]
\centering
\small
\begin{tabular}{ccclllccc} 
\hline
\multicolumn{3}{c}{\textbf{Model}} & \multicolumn{2}{c}{\textbf{Metrics}} & \multicolumn{3}{c}{\textbf{Inference}}\\
dialogue & email & aggregator & \multicolumn{1}{c}{micro-F1} & \multicolumn{1}{c}{acc.} & \multicolumn{1}{c}{params.} & \multicolumn{1}{c}{latency CPU} & \multicolumn{1}{c}{latency GPU}\\ \hline
- & BERT & - & 0.644 & 0.370 & 177.9 & 1.61 & 1.43\\
ToD-BERT & - & - & 0.718 & 0.427 & 109.5 & 1.56 & 1.33\\

ToD-BERT & BERT & Cross Attn. & 0.938 & 0.846 & 288.1 & 2.96 & 2.74\\
ToD-DistilBERT & DistilBERT & Cross Attn. & 0.937 & 0.844 & 202.0 & 2.80 & 2.65\\
ToD-DistilBERT & ALBERT & Cross Attn. & 0.931 & 0.824 & 78.9 & 2.82 & 2.60\\
ToD-DistilBERT & RoBERTa & Cross Attn. & 0.928 & 0.820 & 191.9 & 2.94 & 2.78\\
ToD-DistilBERT & ELECTRA & Cross Attn. & 0.919 & 0.791 & 176.1 & 2.77 & 2.54 \\
ToD-DistilBERT & DeBERTa & Cross Attn. & 0.910 & 0.791 & 205.8 & 3.43 & 2.90 \\
 \hline
\end{tabular}
\caption{\label{table: latency of emtod}
Latency of EMToD framework
}
\end{table*}

\subsection{Deployment}
We compare the existing workflow-based NLU system for intent classification with the EMToD framework in live settings. The scheduling assistant uses the workflow-based system consisting of four ScopeIt and BERT-based models tightly coupled with workflow to track the state of the conversation. The conversation tracking by the workflow allows routing text to the appropriate intent detection model trained on the specific distribution of the data. The existing system beats the pure model baselines in section \ref{sec:baselines}. 

As both methods (workflow and EMToD) are tracking the state of conversation, the comparison between them will indicate how effectively can EMToD utilize conversation history and state for intent detection. So for this experiment, we log and label conversations for 2 weeks in November 2021 between agent and user for the task of scheduling meetings. We run an internal workflow-based system and EMToD on this data and compare the final results. 
From Table \ref{table: deployment results}, we can see that EMToD provides at least 5\% improvement over the internal system. Specifically, EMToD improves intent detection performance by 20\% on conversations with more than 6 turns. This indicates that  EMToD can not only track conversation history and state better than workflow but also utilize the conversation history better to predict user intents.

\subsection{Inference latency of EMToD} \label{sec: Inference time}

We measure the latency of different configurations of EMToD and baselines. The results are shown in Table \ref{table: latency of emtod} where the latency is in seconds and parameters (params) are in Millions. In the table, the inference latency reported is for input tokenization followed by a forward pass of the model with a batch size of 1. The hardware used is a single Nvidia Tesla K80 GPU and single Intel(R) Xeon(R) CPU E5-2690 v3 @ 2.60GHz CPU. The latency numbers reported are an average for 10 runs. The EMToD framework starts after the second entry in Table \ref{table: latency of emtod}. For all EMToD frameworks, we use ScopeIt as user turn summarizer and heuristic summarizer for agent response summary. 

The EMToD configuration with ToD-DistilBERT as dialogue model and DistilBERT as language model provides the best trade-off in terms of performance and inference latency. This particular configuration leads to about 73\% and 85\% increase in CPU and GPU latency over BERT but at the same time increases the intent detection performance by 45\%. Similarly, for ToD-BERT, this configuration leads to about 79\% and 99\% increase in CPU and GPU latency but at the same time increases the intent detection performance by 30\%. Therefore, the increased performance comes with an inference latency tradeoff. In the future, further investigation into distillation methods for EMToD might lead to consolidated and reduced model size and inference latency.

\section{Related work}
Intent detection is one of the core components of NLU and has benefited a lot from pre-trained language models \cite{devlin-etal-2019-bert, NEURIPS2019_c20bb2d9} and pre-trained dialogue models \cite{wu-etal-2020-tod, zhang-etal-2020-dialogpt}. Jointly modeling intent detection with slot filling has shown to improve performance \cite{6639291, Liu2016AttentionBasedRN, e-etal-2019-novel}. In order to reduce labeling efforts, researchers have developed mining intents from past conversations \cite{chatterjee-sengupta-2020-intent, shi-etal-2018-auto}. For email conversations, the downstream task performance including intent detection can be improved by filtering out irrelevant sentences to the assistant's action \cite{patra-etal-2020-scopeit} and then further improving it in an online setting \cite{zhou2021narle}. There has also been a recent focus on identifying tasks from the emails \cite{mukherjee-etal-2020-smart, diwanji-etal-2020-lin} and using user actions in emails as a weak label for improving intent detection models \cite{10.1145/3397271.3401121}. Incorporating context has been shown to improve performance for intent detection and other tasks in both email and short conversation settings \cite{6639291, 10.1145/3331184.3331260, trajanovski-etal-2021-text}. In this paper, we utilize the conversational context and transfer learning to improve intent detection for email conversations.
\section{Conclusion}
In this paper, we propose EMToD, a simple and effective transfer learning framework (EMToD) that allows the latest development in dialogue models to be adapted for email conversations. We show that the EMToD framework results in 30\% and 45\% improvement over pre-trained task-oriented dialogue and language models. We analyze the contribution of each component of EMToD towards performance improvement and compare the framework's performance over traditional workflow-based NLU systems in deployment settings. The modular nature of EMToD allows easy translation of future improvements in both pre-trained language and task-oriented dialogue models toward building email-based assistants. In the future, we plan to investigate the utility of EMToD for dialogue state tracking and dialogue act prediction in email conversations.

\bibliography{anthology,custom}
\bibliographystyle{acl_natbib}
\clearpage

\end{document}